\documentclass[a4paper, 12pt]{article}
\usepackage[utf8]{inputenc}
\usepackage[T1]{fontenc}
\usepackage{latexsym}

\usepackage{graphicx}
\usepackage{amsmath}
\usepackage{amsfonts}
\usepackage{amssymb}

\usepackage{array}
\usepackage{dsfont}
\usepackage{multirow}
\usepackage{lineno}
\usepackage{setspace}
\usepackage{float}
\usepackage{color}
\usepackage{times}

\usepackage{hyperref}
\newcommand{\bc}{\begin{center}}
\newcommand{\ec}{\end{center}}
\newcommand{\beq}{\begin{equation}}
\newcommand{\eeq}{\end{equation}}
\newcommand{\beqa}{\begin{eqnarray}}
\newcommand{\eeqa}{\end{eqnarray}}
\newcommand{\beqan}{\begin{eqnarray*}}
\newcommand{\eeqan}{\end{eqnarray*}}
\newcommand{\bit}{\begin{itemize}}
\newcommand{\eit}{\end{itemize}}

\newcommand{\beqna}{\begin{eqnarray}}
\newcommand{\eeqna}{\end{eqnarray}}

\begin{document}

\title{Brain Tumor Segmentation with Special Emphasis on the Non-Enhancing Brain Tumor Compartment}
\author{$^1$T. Schaffer, $^2$A. Brawanski, $^{1,3}$S. Wein, $^4$A. M. Tomé,  and $^1$E. W. Lang,\\
$^1$CIML Group, Biophysics, University of Regensburg, \\ 93040 Regensburg, Germany\\
$^2$Department of Neurosurgery , University Hospital Regensburg,\\ 93042 Regensburg, Germany  \\
$^3$Department of Biomedical Imaging, University Hospital Regensburg, \\ 93042 Regensburg, Germany \\
$^4$DETI, IEETA, Universidade de Aveiro, 3810-195 Aveiro, Portugal \\
Email: {\em alexander.brawanski@ukr.de}, {\em elmar.lang@ur.de}
}

\maketitle

%%%%%%%%%%%%%%%%%%%%%%%%%%%%%%%%%%%%%%%%%%%%%%%%%%%%%%%%%%%%%%%%%%%%%%%%%%%%%%%%%%%
\section{Introduction}

Glioblastomas are the most malignant type of brain tumor of the adult age and the overall prognosis is still poor. The overall survival time is around 18 months despite the application of diversified treatment options, albeit there have been advances in treatment by analysis of molecular subtypes and an adapted therapeutic regime. As treatment options surgery, radiotherapy and chemotherapy are available. Treatment planning can be based on either X-ray computed tomography (CT), single-photon emission computed tomography (SPECT), positron emission tomography (PET) and, most frequently, on magnetic resonance imaging (MRI). Especially MRI, as a non-invasive technique, is the most promising image modality to localize the tumor in the brain and reveal its structural details. Concerning subsequent image processing, four multi-parametric MR image modalities are applied frequently like T1-weighted (T1), T2-weighted (T2), T1-weighted with gadolinium contrast enhancement (T1C) and T2-fluid attenuated inversion recovery (FLAIR). All of them are co-registered to a common anatomical template \cite{Rohlfing2010}, resampled to a uniform isotropic spatial resolution of $1\ mm^3$ and skull-stripped. Recently more advanced diffusion and perfusion MRI procedures such as diffusion tensor imaging (DTI), diffusional kurtosis imaging (DKI), dynamic susceptibility contrast-enhanced (DSC) perfusion imaging, dynamic contrast - enhanced (DCE) perfusion imaging, arterial spin labeling (ASL), nuclear magnetic resonance (NMR) spectroscopy and amide proton transfer (APT) were applied. These advanced MRI procedures provide additional information to better identify the different characteristics of high grade gliomas (HGGs) \cite{Scola2023}.

Brain tumor segmentation (BTS) is an essential step in glioma diagnosis, but is still a largely open problem despite the many attempts to tackle it \cite{Menze2014}. BTS methods can be classified into generative probabilistic techniques and discriminative approaches. The former try to map input to output distributions based on anatomical knowledge available, while the latter learn directly from data.

From a medical point of view four segments characterize a brain tumor, called necrosis (NCR), contrast-enhancing tumor (ET), non-contrast-enhancing tumor  (NET) and edema (ED). The usual guideline to estimate tumor extension is the T1C gadolinium enhanced MRI. However it is well known that tumor cells extend beyond these areas. On the one hand, these cells are mainly responsible for tumor regrowth, on the other hand they are difficult to identify. One candidate for these cell populations is the NET part in the MRI imaging suite. This NET compartment, however, is ill-defined in terms of clear and limited criteria. There have been efforts to define these areas by radiological criteria, but there is no general consent and the variability of finally defining these areas is high. This was also the obvious reason, why the direct segmentation of these areas was abandoned in segmentation trials. Thus, until 2015/16 the brain tumor segmentation (Brats) competitions included the NET segment in their data, but recent BTS competitions \cite{BraTS2022} prefer to segment only three of those compartments, while the NET compartment is combined with one of the others \cite{Liu2023}. This non-enhancing peritumoral area is commonly defined as the hyperintense region in T2 and FLAIR images surrounding a contrast-enhancing brain tumor compartment. In MRI recordings, the NET compartment is typically found as a border region between the edema and the enhancing tumor compartment. The NET compartment seems associated with various pathological processes \cite{Lemee2015}, \cite{Villanueva-Meyer2017b}, including vasogenic edema and infiltrative edema. While the former is associated with brain metastases, the latter can be related to diffuse gliomas \cite{Artzi2014}, \cite{Villanueva-Meyer2017}, \cite{Lah2020}.  Additionally, the analysis of the peritumoral area helps in assessing the presence of residual or recurrent tumors and in diagnosing tumor progression. Especially in HGGs the tumor infiltration of neoplastic cells is often beyond the contrast-enhanced tumor compartment into the non-enhancing compartment of the tumor \cite{Scola2023}. The infiltrative edema is known to be responsible for HGG recurrence and tumoral spreading after apparently complete surgical resection \cite{Martin-Noguerol2021}. In fact there is also convincing evidence of the importance of the NET compartment \cite{Brenner2022} for the long term survival of the patients \cite{Schoenegger2009}, \cite{Lasocki2019}, \cite{Pasquini2021}.  Advanced MRI investigations also suggest NET being a promising tool for distinguishing HGGs from brain metastases and predicting clinical outcome and response to surgery and chemo-irradiation. A timely recent survey considering MRI investigations focusing on the non-enhancing peritumoral tumor compartment is given by \cite{Scola2023}. Hence, accurate segmentation of all four tumor compartments can offer the basis for an improved quantitative image analysis.

The aim of our study thus was to construct a way to extract the non-enhancing tumor compartment by utilizing available information about this segment in publicly available datasets. The newly proposed method provides a reliable way to extract NET compartments thereby offering a more detailed and informative analysis of brain tumor MRI datasets. Furthermore the new method offers two advantages over existing models: On the one hand, it automatically generates resolution enhanced segmentation masks helping radiologists and clinicians to better grasp the shape and extent of the tumor and its compartments. On the other hand, in addition to the traditional segments, it also predicts the important NET - compartments, which are hard to locate on the conventional radiology records. In summary, the presented study is based on a modified U-Net architecture with an extension to enhance the spatial resolution of the reconstructed MRI images. The method outperforms various variants of standard and advanced U-Nets with different filter blocks. Various evaluation metrics favorably compare to published results of the MICCAI competitions.

%%%%%%%%%%%%%%%%%%%%%%%%%%%%%%%%%%%%%%%
%%%%%%%%%%%%%%%%%%%%%%%%%%%%%%%%%%%%%%%
%%%%%%%%%%%%%%%%%%%%%%%%%%%%%%%%%%%%%%%
\section{State-of-the-Art}

While earlier surveys focused on classic BTS methods \cite{Gordillo2013}, \cite{Liu2014}, \cite{Kapoor2017}, \cite{Hameurlaine2019}  the last years (since 2014 roughly) have witnessed an increasing interest in BTS techniques based on deep learning \cite{Akkus2017}, \cite{Ghaffari2019},  \cite{Biratu2021}, \cite{Magadza2021}, \cite{Liu2023}.  MRI techniques are the major source of information in clinical practice. But low contrast images follow from weak MRI signal intensities due to small magnetic susceptibility differences between certain tumor compartments. Especially at the borders of edemas differences in magnetic properties to neighboring tissues are small rendering and edemas are hard to delineate precisely. The same is true for the non-enhancing compartment (NET). This peritumoral area is commonly defined as the hyperintense region in T2-weighted and FLAIR images surrounding a brain tumor. An additional complication arises from label imbalanced datasets causing the feature extraction process to be dominated by the larger brain tumor compartments. Because manual segmentation is time-consuming, annotation biased and error-prone subject to large inter- and intra-observer variability, recent efforts have concentrated on automatic segmentation techniques based on machine learning.

Early approaches focused on classical machine learning methods employing hand-crafted features but were constrained by computational resource limitations \cite{Zhu1997}, \cite{Kaus1999}, \cite{Corso2008}, \cite{Menze2010}. Furthermore, hand-crafted features introduce a strong bias due to inevitable prior knowledge that needs to be available but cannot be generalized easily. Also only the entire tumor has been segmented mostly, thus lacking any insight into the internal structure with the spatial distribution of the four compartments. Fully automated BTS methods deployed, beneath others, Bayesian and Markov random field methods \cite{Prastawa2004}, \cite{Bauer2010}, atlas-based registration \cite{Prastawa2003}, \cite{Cuadra2004}, \cite{Xue2006}, statistical  models of tumor deformation\cite{Mohamed2006}, binary support vector machine (SVM) classifiers \cite{Bauer2011}, \cite{Zhang2011} and, most notably, random forest (RF) multi-class classifiers \cite{Criminisi2013}, \cite{Geremia2013}, \cite{Tustison2015}, \cite{Maier2015}, \cite{Pinto2015}. Random forests construct an ensemble of decision trees during training and the class collecting most votes is selected during testing. RFs utilize large numbers of different features in their high-dimensional input vectors and can naturally handle multiple classes simultaneously for classifications. Typically probabilistic maps are generated from the RF as context-aware spatial priors, which subsequently may be refined by Markov Random Field (MRF) regularization \cite{Tustison2015}. Further optimization concerns the relation between the out-of-bag (OOB) error and the Dice coefficient, though the number of classes is limited to whole tumor and tumor core without edema \cite{Lefkovitz2017}. Also RFs have been combined with level set methods, where in a first phase a voxelwise classification into four categories was performed with an RF, and in a second phase accurate tumor boundaries were established via active contours \cite{Chan2001} based on level set methods \cite{Lefkovitz2018}.

With the advent of AlexNet \cite{Krizhevsky2012} the area of deep learning soon entered the field of BTS \cite{Zikic2014}, \cite{Lyksborg2015}, \cite{Pereira2016}, \cite{Zhao2016}, \cite{Havaei2017}, \cite{Zhuge2017} with customized deep convolutional networks (DCNN) \cite{Long2015}. The latter automatically learn a set of complex features from the training set. Subsequently they can predict the class of a pixel/voxel by processing a usually small patch centered at that pixel/voxel and primarily focus on local features. The general architecture encompasses stacked modules of multilayer perceptrons (MLPs) and pooling layers, occasionally enhanced by normalization layers, dropout regularization and skip connections \cite{Goodfellow2016}. The final layer depends on the actual application, and in case of a classification task is often represented by a softmax layer. Furthermore, data augmentation is frequently applied to offset the disadvantage of only small training sets, which is often met in biomedical image analysis. Hence convolutional neural networks quickly became the standard for automatic feature extraction using deep learning techniques \cite{Kamnitsas2016-BraTS2015}, \cite{Kamnitsas2017}, \cite{Le2020}, \cite{Sua2020}. A major breakthrough happened with the introduction of the U-Net architecture by \cite{Ronneberger2015}, \cite{Cicek2016}, \cite{Yin2022} or its related V-Net \cite{Milletari2016}. U-Nets are based on fully convolutional networks (FCN) and follow an encoder - decoder design. The encoding path consists of stacked modules of representation layers followed by a pooling layer. While during feature extraction the number of filters is increased, every pooling layer reduced the spatial resolution but concomitantly increases the related receptive field in the input layer thus allowing for the extraction of ever more complex features. The decoding path inverts these operations in every module and finally outputs a segmented spatial map rather than only classification scores. Decoding is supported by skip connections, which at every level of spatial resolution copies and concatenates the feature maps from the encoding to the corresponding decoding stage. As convolutional operations are intrinsically local, U-Nets, however, have difficulties to properly represent long-range dependencies. Training is commonly based on a cross-entropy loss function, but with the V-Net a new loss function based on the Dice coefficient was introduced, which turned out to be advantageous with imbalanced datasets as they often happen with biomedical images. Recently a deep learning-based segmentation method, called nnU-Net \cite{Isensee2021}, has been proposed that automatically configures itself, including preprocessing, network architecture, training and post-processing for any new task. It provides a from-the-shelf toolbox for semantic image segmentation tasks \cite{Liu2018}. Numerous other modifications exist as, for example, the semantic segmentation method for 3D brain tumor segmentation proposed by \cite{Myronenko2018}, the AttU-Net with its integrated attention gate \cite{Oktay2018}, the R2U-Net, an integration of a recurrent neural network and a  ResNet into the original U-Net \cite{Alom2018}, the Ternausnet \cite{Iglovikov2018}, the boundary - aware dual U-Net \cite{Li2019}, the UNet++ \cite{Zhou2020}, the CMM-Net \cite{Al-Masni2021}, the MSU-Net with its multi-scale blocks containing different receptive field sizes \cite{Su2021}, the transformer-based TransUNet \cite{Chen2021}, which combines the transformer concept with the U-Net encoder-decoder principle and deploys sequence-to-sequence learning and a self-attention mechanism based on conditional random fields \cite{Cuong2014} for global feature extraction \cite{Dosovitskiy2020}, the  FANet \cite{Tomar2022}, the BTS U-net \cite{Aumenete-Maestro2022} or the tokenized convolutional MLP called UNeXt \cite{Valanarasu2022}. Most recently the Segment-Anything project \cite{Kirilov2023} has been launched (https://segment-anything.com) providing  a new task, model and dataset (over 1 billion masks and 11M images) for image segmentation.

Numerous applications of CNNs to BTS of pre-operative MRI scans have been presented in the BraTS challenges \cite{Menze2010} of the MICCAI conferences since 2012 \cite{Zikic2014}, \cite{Urban2014}, \cite{Davy2014}, \cite{Menze2015}, \cite{Rao2015}, \cite{Havaei2017}, \cite{Roth2018}, \cite{Myronenko2018}, \cite{Bakas2019}. Early BraTS challenges (2012 - 2016) dealt with four heterogeneous histological subregions in general, named necrosis (NEC/NCR - label 1), active or enhancing tumor (AT or ET - label 4), non-enhancing tumor (NET - label 3) and peritumoral edema (ED - label 2). Alternatively, subregions were also labeled as whole tumor (WT), tumor core (TC), enhancing tumor (ET) and edema (ED) \cite{Menze2015}. The AT/ET areas show hyper-intensity in T1C when compared to T1. The TC compartment describes the bulk of the tumor, which is what is typically resected and entails the AT/ET, as well as the necrotic and the non-enhancing parts of the tumor. The necrotic (NCR) and the non-enhancing (NET) tumor compartments typically show hypo-intensity in the T1C modality when compared to T1. The WT denotes the complete extent of the tumor and entails the TC and the peritumoral edematous tissue (ED), which is typically depicted by a hyper-intense signal in T2-FLAIR \cite{Bakas2019}. Note that these subregions do not represent well defined biological entities, but are rather image based. Later applications (starting from 2017 on) only considered three classes, variously called whole tumor (WT - labels 1 to 4), gross tumor core (TC - labels 1,3 and 4) and active tumor (AT - labels 3 and 4), by merging the difficult to segment non-enhancing tumor compartment with one of the other compartments (see section Data for details). Since 2017, BraTS challenges also considered the prediction of overall survival of patients diagnosed with primary glioblastoma. Also since the 2017 BraTS challenge, a validation set was included beneath the training and testing datasets. The winning results of the BraTS challenges 2017 to 2022 are listed in an a supplementary material for the convenience of the reader.

%%%%%%%%%%%%%%%%%%%%%%%%%%%%%%%%%%%%%%%
%%%%%%%%%%%%%%%%%%%%%%%%%%%%%%%%%%%%%%%
%%%%%%%%%%%%%%%%%%%%%%%%%%%%%%%%%%%%%%%
\section{Datasets}

Glioblastomas can be subdivided into four grades depending on microscopic images and tumor behavior. Low-Grade-Gliomas (LGG - Grade I and II) are close to benign and slowly growing, while High-Grade-Gliomas (HGG - Grade III and IV) are cancerous and very aggressive. HGG and LGG datasets from the BraTS 2015, 2018, 2021 and 2022 challenges  were employed in this study for training, validation and testing. The data was recorded on a variety of scanners with magnetic inductions - often sloppily called field strength - of either $1.5\ T$ or $3\ T$. All datasets encompassed four MRI modalities: T1-weighted, contrast-enhanced T1-weighted, T2-weighted and T2-FLAIR images. All images were co-registered to the same anatomical reference based on a mutual information similarity measure. Also all images were skull-stripped and interpolated to a voxel resolution of $1 mm^3$. Annotation of all datasets was performed by experts and all data went through a review and approval process.

The BraTS 2015 dataset \cite{Menze2015} received labels of the following categories: background (0), necrosis (1), edema (2), non-enhancing tumor (3) and enhancing tumor (4). Compared to the BraTS 2018 and BraTS 2021 datasets this is the only one which provides dedicated labels for the NET compartment. However, the BraTS challenges do not use the labels as given in the datasets but hierarchical combinations of them resulting in the following evaluation segments: active tumor (AT), tumor core (TC) and whole tumor (WT). These evaluation segments are identical in all considered BraTS challenges, but the provided data labels as well as the relation between them and the evaluation segments vary between the different BraTS datasets.

In the BraTS 2015 dataset, the data labels NCR, ET, NET, ED and the evaluation segments AT, TC, WT were linked as follows:

\bit 
\item 
$AT \leftrightarrow NCR + ET$
\item 
$TC \leftrightarrow NCR + ET + NET$
\item 
$WT \leftrightarrow NCR + ET + NET + ED$
\eit 

The BraTS 2018 dataset \cite{Bakas2019} provides a set of $285$ multimodal (T1, T1C, T2, FLAIR) MRI records of human brains, each containing scans of a brain with HGG and LGG tumors. In addition, the dataset also contains overall survival data for each of the records. The data labels are composed as follows: background (0), NET plus NCR (1), ED (2), empty (3) and ET (4). The announcement paper \cite{Bakas2019} does not state any reason for this change, also any obvious medical and technical reason is missing. Finally the evaluation segments are obtained as:

\bit 
\item 
$AT \leftrightarrow ET$
\item 
$TC \leftrightarrow (NCR + NET) + ET $
\item 
$WT \leftrightarrow (NCR + NET) + ET + ED$
\eit 

The BraTS 2021 challenge \cite{Baid2021} provides a total of $2040$ records, where $1251$ of them come as training data together with labels. Again data labels changed resulting in background (0), NCR (1) ED plus NET (2), empty (3) and ET (4). Also the evaluation regions are varied again, whereby the segment AT has been dropped and the segment ET is used to denominate the smallest of the hierarchical parts. The TC compartment is now a combination of the center necrotic part and the surrounding enhancing region, but without the non-enhancing compartment as it was used in both previous datasets. Instead the NET compartment is now connected to ED, hence only count in case of the WT segment, i.e. we now have

\bit 
\item 
$ET \leftrightarrow ET$
\item 
$TC \leftrightarrow NCR + ET $
\item 
$WT \leftrightarrow NCR + ET + (ED + NET)$
\eit 

Finally the training and validation data of the BraTS 2022 dataset are identical to the BraTS 2021 dataset. Only the testing dataset has been updated with additional MRI scans \cite{Bionetworks2022}.
%%%%%%%%%%%%%%%%%%%%%%%%
%%%%%%%%%%%%%%%%%%%%%%%%
%%%%%%%%%%%%%%%%%%%%%%%%
\section{Methods}

\subsection{Image preprocessing}

As a pre-processing step, the multi-channel input records have been center-cropped to size $80 \times 160 \times 128 \times 4$ (axial, coronal, sagittal, channel). Cropping had nearly no effect regarding image data loss, since a wide outer area of the records was empty. Working with the original full record size of $155 \times 240 \times 240 \times 4$ has not been possible due to GPU memory limitations.  The dataset labels NCR, ET and ED were transformed to the evaluation segment labels ET, TC and WT as described above for the different BraTS datasets.

Learning was achieved by optimizing a soft Dice - S\o{}rensen loss function \cite{Milletari2016} in accord with a Dice - S\o{}rensen evaluation metric. In case of multi-class segmentation tasks, the soft Dice loss has been computed for each class separately and then averaged over all classes. With this loss function, the network produced probabilistic segmentations, which finally could be thresholded to yield binary segmentation results. For further evaluation the familiar metrics Intersection over Union (IoU) and Hausdorff Distance (HD) have been applied additionally.

%%%%%%%%%%%%%%%%%%%%%%%%%%%
%%%%%%%%%%%%%%%%%%%%%%%%%%%
\subsection{Reference segmentation model}

\subsubsection{3D U-Net with general filter blocks}

The proposed segmentation model resembles a canonical U-Net \cite{Ronneberger2015}, a slightly amended version of which served as a reference model. The principle architecture resembles an encoder-decoder design with an encoding path followed by a decoding path, each going through four levels of spatial resolution and each of the three spatially related top levels having a skip connection. The structure within the single layers is repeated with each level having a different number of filters and a different spatial resolution. \\

The multi-channel input records have been center-cropped to size $80 \times 160 \times 128 \times 4$ (axial, coronal, sagittal, channel). After the input layer follows an initial 3D-convolutional layer with $32$ filters of kernel size $3 \times 3 \times 3$ and a subsequent spatial dropout layer with dropout rate $0.2$ \cite{Cai2019}.\\

The top layer of the encoding part furthermore contains a combination of two general filter blocks. The filter blocks are characterized by a structure consisting of normalization, activation, convolution and residual connections. This double filter block structure repeats throughout the model. The  internal composition of filter blocks is kept variable as discussed later. \\

The result of the two sequential filter blocks was spatially down-sampled to half size using $3 \times 3 \times 3$ convolutions with stride 2. Note that stride 2 convolutions are capable of carrying parameters for the down-sampling operation, whereas max-pooling is a fixed operation. Thus, potentially, more complex and task-specific patterns can be learned, which may improve the performance. Also stride 2 convolutions can be computationally more efficient than max-pooling combined with spatially invariant convolutions \cite{Ayachi2020}.\\

In the subsequent decoding path, the upsampling is performed by applying transposed 3D convolutions of stride 2. After each of the upsampling steps, the result was summed, rather than concatenated, with the skip connection path connecting from the encoding side. After reaching back to the original spatial shape another two residual filter blocks were applied, followed by a 1 × 1 × 1 convolution. The final convolution used three filters, yielding three output masks while deploying a sigmoidal activation function. These outputs represented the three prediction probability masks for enhancing tumor (ET), tumor core (TC) and whole tumor (WT), respectively. This completed a typical U-Net architecture taken as reference model during model evaluation.

%%%%%%%%%%%%%%%%%%%%%%%%%%%%%%
\subsubsection{Filter block structures}

\begin{figure}[!htp]
  \centering
  \includegraphics[width=0.9\textwidth]{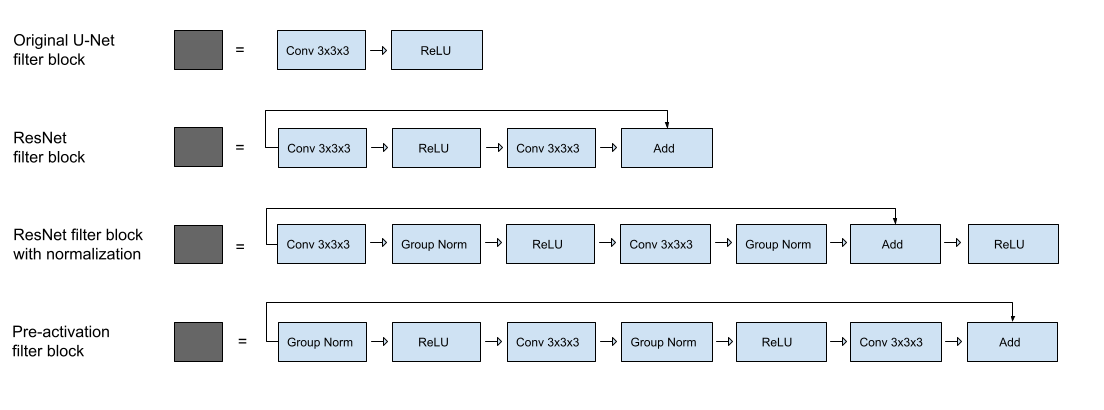}
  \caption{Different types of residual filter blocks. Each type can be used in the above U-Net architecture}
  \label{fig:fig6}
\end{figure}

Historically, there are basically three  main iterations of filter blocks which have been used. In the original U-Net paper \cite{Ronneberger2015} plain convolutions with ReLU activation were used as shown in the top row of Figure \ref{fig:fig6}, while a later model deployed ResNet like filter blocks. Their characteristic structure contains a combination of two subsequent convolutions with an intermediate ReLU activation and an additive residual connection bypassing the convolution operations. Yet further model modifications included normalization blocks, where commonly an activation layer followed the convolution layer. The residual connection may be added before or after the final activation operation. Here the version with activation after the residual connections has been implemented and is shown in the third row of Figure  \ref{fig:fig6}. Batch normalization has been replaced by group normalization, since the model had to be trained with batch size 1. An example of this filter block structure is shown at the bottom row of Figure \ref{fig:fig6}. Note that the normalization and activation layers were placed before the convolution layers \cite{Myronenko2018}, hence the name pre-activation U-Net.

These modifications result in four different base models, which have been trained and evaluated on the BraTS data for comparing their performance. The pre-activation filter blocks were selected for the further extraction process of the NET - segment. A comparison of learning curves during training and testing, deploying the BraTS 2018 and BraTS 2021 datasets, is provided in the supplement.

\subsection{An upscaling PAU-Net}

\begin{figure}[!htbp]
  \centering
  \includegraphics[width=0.8\textwidth]{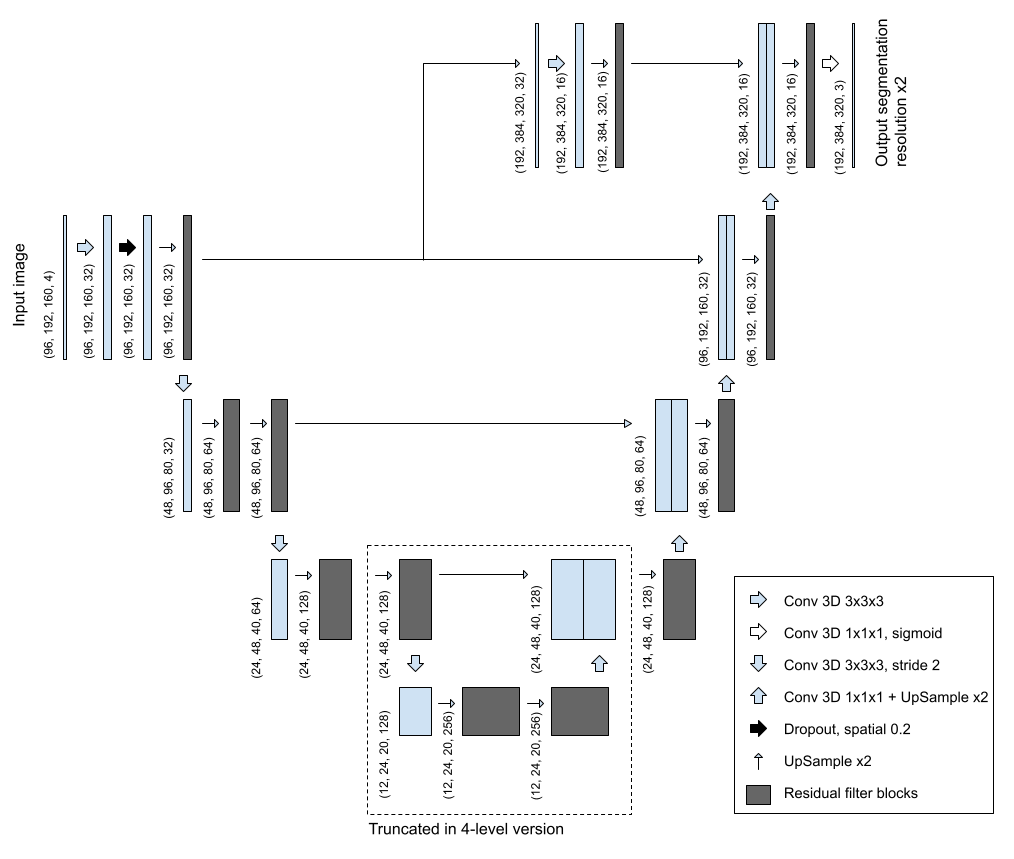}
  \caption{Two variants of the new up-scaling PAU-Net.}
  \label{fig:fig8}
\end{figure}

Figure \ref{fig:fig8} shows two versions of the upscaling pre-activation U-Net (PAU-Net) model which has been deployed to run the segmentations on the BraTS 2018 and 2021 datasets. The new model extends the best performing base model by the addition of a further upscaling branch on the decoder side, which exceeds the spatial input resolution by a factor of 2. This increases the output segmentation mask resolution and therefore serves the detection of small scale structures, as they are characteristic for NET - compartments. Additionally there is a filtered skip path connecting the new high resolution level with the first of the original skip connections. This is a rare design choice \cite{Myronenko2018}, \cite{Myronenko2020}, since U-Nets traditionally come with a rather symmetric design. The ground-truth masks have been adapted by upscaling them to double size. Up-scaling the binary masks was achieved by repeating the given binary values, no interpolation was used. The Dice - S\o{}rensen coefficient metric as well as the Soft-Dice loss function were accordingly adapted to the increased size of the prediction and ground truth maps.

%%%%%%%%%%%%%%%%%%%%%%%%%%%%%%%%%%%%%%%%%%%%%%%%%%%%%%%%
%%%%%%%%%%%%%%%%%%%%%%%%%%%%%%%%%%%%%%%%%%%%%%%%%%%%%%%%
\subsection{Extraction of NET compartments from BraTS 2018 data}

As described above, the BraTS datasets after 2015 \cite{Menze2015} did not come with dedicated segmentation masks for the NET - compartments \cite{Bakas2019, Baid2021}. Rather they were merged with other tumor segments to create hierarchical labels denoting enhancing tumor (\emph{ET}), tumor core (\emph{TC}) and whole tumor (\emph{WT}).  These global labels, however, differed between the BraTS 2018 and BraTS 2021 data. But using a model trained on BraTS 2021 data, we could decompose fused NCR masks from the BraTS 2018 dataset \cite{Bakas2019} and isolate hidden NET regions. Together with the application of morphological filters, this decomposition provided a subset of records with isolated NET masks, which served for further model training.

Figure \ref{fig:unet_upscale_training_results} shows a sample slice from the BraTS 2021 dataset. The first segmentation mask shows the data provided by BraTS 2021, while the second segmentation is extended by a \emph{NET} label, predicted by the PAU-Net.

The BraTS 2018 challenge paper \cite{Bakas2019} states that the Necrosis segment \emph{NCR} is labeled together with the Non-Enhancing segment \emph{NET}. Therefore the \emph{NET} segment is implicitly included in the BraTS 2018 \emph{NCR} labels. This may have it's reason in both regions being represented as hypo-intense parts in the contrast enhancing records T1C. Nevertheless in most cases NCR and NET are generically located in spatially well separable regions. Necrosis \emph{NCR} is typically found in center tumor parts, surrounded by the Enhancing Tumor \emph{ET} region, and forming the Tumor Core \emph{TC} together with it. Outside of this \emph{TC} region, in it's bordering area and reaching into the surrounding Edema \emph{ED} domain, the Non-Enhancing \emph{NET} part can be found. In many cases Necrosis \emph{NCR} and Non-Enhancing Tumor \emph{NET} have only few or no touching or intersecting points. By a process of predicting, filtering and subtracting the plain Necrosis regions using the trained BraTS 2021 model, it is widely possible to separate the two regions.

The basic prediction and classification steps were the following ($X^{pred}$ denotes prediction from a PAU-Net model trained on BraTS 2021 data, $X^{gt}$ denotes ground truth labels from BraTS 2018 data) :
\bit
\item
For a given BraTS 2021 record do the following:
   \bit
       \item
       Train a PAU-Net end-to-end on the BraTS 2021 dataset
       \item
       Predict the labels $ET_{21}^{pred}$, $WT_{21}^{pred}$ and $TC_{21}^{pred} = NCR^{pred}_{21} \;\cup\;
       ET_{21}^{pred}$.
       \item
       Compute $NCR^{pred}_{21} = TC_{21}^{pred} \setminus ET_{21}^{pred}$ labels using the trained PAU-Net.
   \eit
\item
For any given BraTS 2018 record do the following (note that
$NCR^{gt}_{18} \neq NCR^{gt}_{21}$)
   \bit
         \item
         Read given $NCR_{18}^{gt} = NCR^{gt}_{21} \;\cup\; NET$.
         \item
         Predict $NCR^{pred}_{21}$ using the trained PAU-Net.
         \item
         The NET-region can then be extracted as $NET = NCR^{gt}_{18} \setminus NCR^{pred}_{21}$
         or even as $NET = NCR^{gt}_{18} \setminus TC^{pred}_{21}$ given that the NET-part does
         not reach into the ET-region.
   \eit
\item
    Reassign pixel labels to $NET$ if pixel is classified as $NCR^{gt}_{18}$ but not
    $TC^{pred}_{21}$. Also Edema $ED$, Enhancing Tumor $ET$ and background pixels will be
    changed to $ET$ if classified as such.
\eit

For improving the results of the above process, morphological filters were applied to smooth the masks and to reduce noise effects.  A detailed list of the operation steps and the applied filters is listed in %Table \ref{table:morpholocial_filters}
the supplementary material. The major advantages of using morphological filters are:

\begin{itemize}
    \item
    Reduction of artifacts resulting from subtraction of masks.
    \item
    Elimination of small (single pixel sized) border noise, which was present in some of the given BraTS 2018 ground truth masks.
    \item
    Filling of small gap regions in border areas.
\end{itemize}

\begin{figure}[!htbp]
  \centering
  \includegraphics[width=300pt]{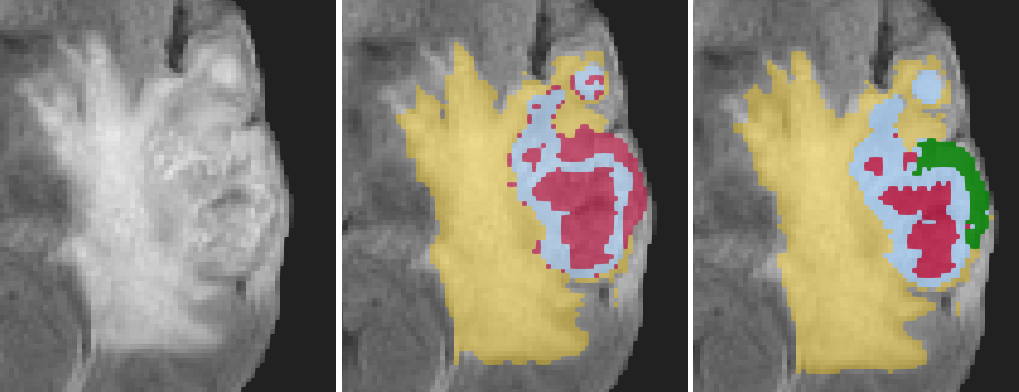}\\
  \vspace{10pt}
  \includegraphics[width=300pt]{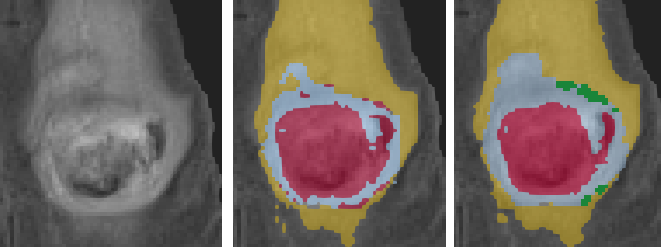}\\
  \vspace{10pt}
  \includegraphics[width=300pt]{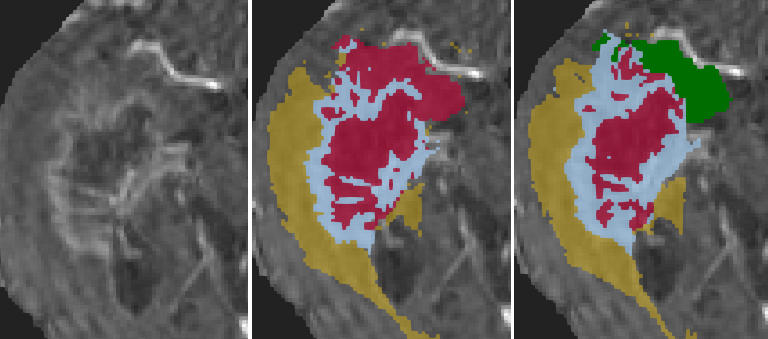}\\
  \vspace{10pt}
  \includegraphics[width=300pt]{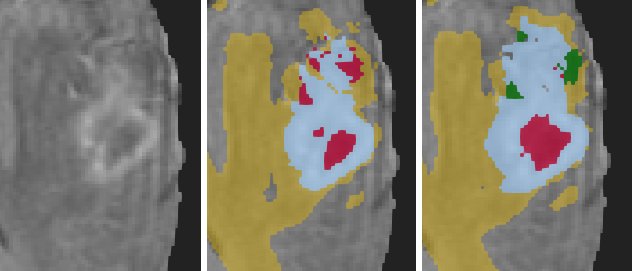}\\
  \caption{Records from the BraTS 2018 dataset \cite{BraTS2018} named CBICA-ANI-1, CBICA-AXO-1, CBICA-AQQ-1, TCIA03-419 (top to bottom). Left: Flair (CBICA-ANI-1, CBICA-AXO-1) or T1C (CBICA-AQQ-1, TCIA03-419) slice. Center: original BraTS 2018 segmentation mask (ED yellow, NCR+NET red, ET blue). Right: NCR (red) and NET (green) separated and filtered.}
  \label{fig:brats2018_processed_net_samples}
\end{figure}

The sample images in Figure \ref{fig:brats2018_processed_net_samples} show slices of BraTS 2018 records which have been processed by the described operations.

Having obtained isolated \emph{NET} masks, the BraTS 2018 dataset was analysed with respect to the total volumes $V_{NET}$ of the isolated {\em NET} compartments. Therefore the total volume of the \emph{NET} mask was measured in each record and the data was partitioned into three groups using Gaussian Mixture Model Clustering. Details of the procedure are given in the supplementary material.
%%%%%%%%%%%%%%%%%%%%%%%%%%%%%%%%%%%%%%%%%%%%%%%%%%%%%%
%%%%%%%%%%%%%%%%%%%%%%%%%%%%%%%%%%%%%%%%%%%%%%%%%%%%%%
%%%%%%%%%%%%%%%%%%%%%%%%%%%%%%%%%%%%%%%%%%%%%%%%%%%%%%
\section{Results}

The four presented filter block structures have been used in the basic U-Net model and have been compared against each other by training and evaluating the derived models on the BraTS 2018 and BraTS 2021 data. Corresponding results are compiled in the supplementary material.

%%%%%%%%%%%%%%%%%%%%%%%%%%%%%%%%%%%%%%%
\subsection{Training of the upscaling PAU-Net model}

The new model has been trained on two different architectures. The first one extended the PAU-Net with the described resolution enhancing extension. Because the last encoding layer carried a very high number of parameters,  it has been eliminated in the second architecture, which brought the model back to a four level design. This was intended to support regularization and allowed larger spatial inputs within limited GPU memory constraints (see Table \ref{table:preact_level_comparsion}).

\begin{table}[h!]
\caption{Model comparison regarding number of parameters, input size and training speed (measured on a Tesla A100-SXM4-40GB GPU)}
    \label{table:preact_level_comparsion}
\begin{footnotesize}
  \begin{center}
    \begin{tabular}{l|l|l|l} % use S for aligned numbers
      \emph{Model} & \emph{Parameters}& \emph{Input size}& \emph{Training speed}\\\hline
      Upscaling PAU-Net 5-level & 13.15M & $80 \times 160 \times 128 (1.64M)$ & 696 ms/step \\\hline
      Upscaling PAU-Net 4-level & 2.87M & $96 \times 192 \times 160 (2.95M)$ & 1096 ms/step \\
    \end{tabular}
  \end{center}
\end{footnotesize}
\end{table}

\begin{figure}[!htbp]
  \centering
  \includegraphics[width=\textwidth]{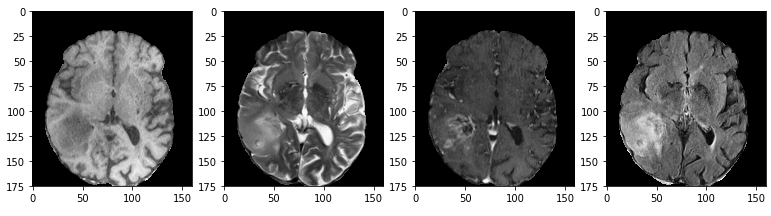} \\
  \includegraphics[width=0.5\textwidth]{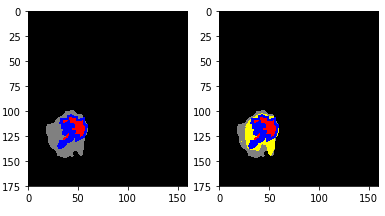}
  \caption{Sample record from BraTS 2021 dataset \cite{BraTS2021}. Top row from left to right: T1, T2, T1C, Flair; Bottom row left: BraTS 2021 tumor segmentation data, right: tumor segmentation data extended with NET region (yellow)}
  \label{fig:unet_upscale_training_results}
\end{figure}

Training results are illustrated in Figure \ref{fig:unet_upscale_training_results} and summarized in Table \ref{table:unet_upscale_training_results}. The 4-level version achieved competitive testing data results for BraTS 2018 and even higher results on BraTS 2021 testing data. The test performance difference of both models on BraTS 2018 in classes TC and WT was very small, while the 5-level version was dominating on ET slightly. On the Brats 2021 test data, the 4-level model’s performance was clearly superior. In total, on the test data, the 4-level version showed better performance than the 5-level variant. Therefore the presented upscaling 4-level PAU-Net has been chosen for all subsequent evaluations.

\begin{table}[h!]
\caption{Upscaling PAU-Net results, Dice-S\o{}rensen coefficients for Enhancing Tumor (ET), Tumor Core (TC) and Whole Tumor (WT) }
    \label{table:unet_upscale_training_results}
\begin{small}
  \begin{center}
    \begin{tabular}{l|c|c||c|c||c|c||c|c} % use S for aligned numbers
       & \multicolumn{2}{c||}{\textbf{BraTS 2018 train}} & \multicolumn{2}{c||}{\textbf{BraTS 2018 test}} & \multicolumn{2}{c||}{\textbf{BraTS 2021 train}} & \multicolumn{2}{c}{\textbf{BraTS 2021 test}}\\

         & 5-level & 4-level & 5-level & 4-level & 5-level & 4-level & 5-level & 4-level\\
      \hline
      ET & 0.7162 & \textbf{0.7179} & \textbf{0.7842} & 0.7802 & 0.8524 & \textbf{0.8566} & 0.8357 & \textbf{0.8464}\\
      TC & \textbf{0.8803} & 0.8646 & 0.8125 & \textbf{0.8230} & \textbf{0.9140} & 0.9121 & 0.8798 & \textbf{0.8841}\\
      WT & \textbf{0.9117} & 0.9010 & \textbf{0.8913} & 0.8912 & \textbf{0.9282} & 0.9232 & 0.9157 & \textbf{0.9186}\\
      \hline
      Mean & \textbf{0.8361} & 0.8278 & 0.8293 & \textbf{0.8315} & \textbf{0.8982} & 0.8973 & 0.8771 & \textbf{0.8830}\\
    \end{tabular}
  \end{center}
\end{small}
\end{table}

\begin{figure}[!htbp]
  \centering
  \includegraphics[width=0.3\textwidth]{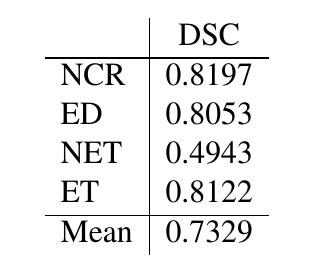} \hspace{10mm}
  \includegraphics[width=0.6\textwidth]{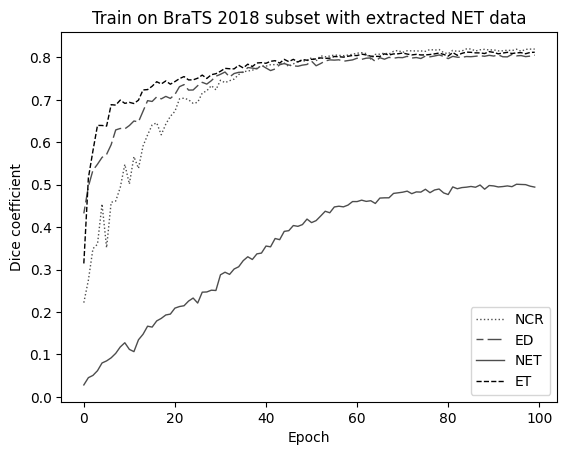}
  \caption{Results for training the PAU-Net on BraTS 2018 subset with extracted NET segments. Training was performed with 100 epochs, 1044ms/step, learning rate $10^{-4}$ with 5\% decay per epoch, training data size 68, batch size 1, GPU Tesla T4. {\em Left:} final DSCs, {\em Right:} Learning curves}
  \label{fig:brats2018_train_subset_net}
\end{figure}

Using the selected subset of medium sized \emph{NET} region data, the 4 - level upscaling PAU-Net model was trained to classify the four segments $ET$, $ED$, plain $NCR$ and the isolated $NET$. Due to having a reduced number of training records together with some degree of inconsistency in the \emph{NET} data, using a classical training/testing split did not return meaningful results. Here, a high-fold cross validation approach would have been needed, which was not executed due to high computational costs. Instead, an early stopping criterion was applied for the $NCR$ Dice coefficient with a minimum absolute increase of $0.005$ being considered an improvement and a patience of 10 epochs. The results are %collected in Table \ref{table:brats2018_train_subset_net} and
illustrated in Figure \ref{fig:brats2018_train_subset_net}. Note that the DSC is preferred for imbalanced datasets because it prevents the model from ignoring the minority classes by focusing on the overlapping regions between the predicted and ground truth masks. The latter is in our case rather a pseudo-ground truth mask with its inherent uncertainty, which explains the rather low DSC. However, visually this might not indicate a weak overlap only. Furthermore, the DSC is strongly biased against single objects, therefore less appropriate for the detection of multiple small structures. \cite{Reineke2023a}.

Having a working model for \emph{NET}, based on the computed subset of isolated \emph{NET} records, the full BraTS 2018 data was again processed using it. As a consequence, the records of group 1, which had few or no \emph{NET} data, became assigned new and additional \emph{NET} predictions, and the third group of data, which originally had a high amount of \emph{NET} pixels, became reduced to smaller \emph{NET} areas. Also the second group of records, which were used as training data, were re-processed to reach a higher level of consistency.
%%%%%%%%%%%%%%%%%%%%%%%%%%%%%%%%%%%%%%%%
%%%%%%%%%%%%%%%%%%%%%%%%%%%%%%%%%%%%%%%%
\subsection{NET volume distributions}

Segmentation outcomes have also been analyzed by estimating the total amount of classified NET-volumes. The latter have been compared with the volumes of the remaining tumor compartments to explore the degree of covariance between the different tumor segments. It was expected that for each of the segments \emph{NCR}, \emph{ED}, \emph{NET} and \emph{ET} volume distributions of similar type could be found when observing the entire dataset. For this analysis, the number of pixels of each of the segments \emph{NCR}, \emph{ED}, \emph{NET} and \emph{ET} was counted throughout the newly extended BraTS 2021 4-label data. The cropped and up-scaled record size in the 4-label dataset was $192 \times 384 \times 320$. Each of the labels defined a subset of pixels, whose volume was counted.
%%%%%%%%%%%%%%%%%%%%%%%%%%%%%%%%%%%%%%%
%%%%%%%%%%%%%%%%%%%%%%%%%%%%%%%%%%%%%%%
%%%%%%%%%%%%%%%%%%%%%%%%%%%%%%%
%%%%%%%%%%%%%%%%%%%%%%%%%%%%%%%
\subsection{Analysis of variance for intensity masks}

For evaluating the newly added \emph{NET} segment and the overall BraTS 2021 4-label dataset towards intensity characteristics of the single regions \emph{NCR}, \emph{ED}, \emph{NET} and \emph{ET} on the different MRI modalities, the data was processed and prepared as follows:

Performing an analysis of variance (ANOVA) of the above region intensities gave the results below, where the extremely low \emph{p}-value suggests that there is a statistically significant difference in the mean intensities across the four categories \emph{NCR}, \emph{ED}, \emph{NET} and \emph{ET}:

\begin{itemize}
    \item \emph{F}-statistic: $44.84$
    \item \emph{p}-value: $1.39 \times 10^{-28}$
\end{itemize}

Since the ANOVA analysis only stated, that there were at least two categories with different means, a post hoc group comparison test had to be performed. This was done by the Tukey Honestly Significant Difference (Tukey HSD) test \cite{Tukey1949}, which compared all possible pairs of means and controlled for the family-wise error rate. Corresponding results are listed in %Tables \ref{table:tukey_t1} to \ref{table:tukey_flair} given in
the supplementary material.
%%%%%%%%%%%%%%%%%%%%%%%%%%%%%%%%%%%%%
%%%%%%%%%%%%%%%%%%%%%%%%%%%%%%%%%%%%%.
\subsection{Training of a unified 4-Label BraTS 2018/2021 Dataset}

The extraction of the NET-regions from the BraTS 2018 data allowed the generation of a new, larger segmentation dataset encompassing four labeled evaluation regions instead of the three original merged versions. With the BraTS 2021 dataset, a cross-validation split, similar to the BraTS 2018 dataset, could be achieved by applying the NET label prediction model, which was deduced from the BraTS 2018 NET reconstruction data. The edema segments could then be cleaned from NET-compartments and the predicted NET segments themselves could be added as a fourth labeled segment. This resulted in two modified BraTS 2018 and BraTS 2021 datasets, providing the new labels NCR, ED, NET and ET. Considering also identical MRI modalities (T1, T2, T1C and Flair records), both modified datasets became compatible and could be unified to one large combined dataset BraTS 2018/21.

The unified BraTS 2018/21 dataset had a total number of $1536$ records, which was divided into a training dataset, holding $1305$ records ($85 \%$ of total), and a test dataset with $231$ records. The structure of the presented upscaling PAU-Net was modified to have an output layer suitable for the five segments ET, TC, WT, NET and TCN.  The training and test data cubes were pre-processed accordingly with a cropped central size of $4 \times 96 \times 192 \times 160$ for each record (channels, axial, coronal, sagittal slices). The Dice-S\o{}rensen coefficient was used as a metric for each of the single segments. The Soft-Dice-Loss function was used as a combined loss function. The batch size was set to one due to high memory demand, and the learning rate was selected to start at $10^{-4}$ with an exponential decay of $5 \%$ per epoch. The GPU accelerator used for training was an NVIDIA A100-SXM4 with $40GB$ RAM, which allowed a training speed of $1177 ms/step$ or approximately  $26 min/epoch$. The training was run for $45$ epochs, which resulted in a total training time of $19.5 h$ for the large prediction model.

The overall BraTS 2021 4-label dataset containing the NET-labeled compartments was further analyzed with respect to intensity characteristics of the individual regions \emph{NCR}, \emph{ED}, \emph{NET} and \emph{ET} on the different MRI modalities, Results are presented in the supplementary material.
%%%%%%%%%%%%%%%%%%%%%%%%%%%%%%%%%%%%%%%%%%%%%%%%
%%%%%%%%%%%%%%%%%%%%%%%%%%%%%%%%%%%%%%%%%%%%%%%%
%%%%%%%%%%%%%%%%%%%%%%%%%%%%%%%%%%%%%%%%%%%%%%%%
\section{Discussion}

\subsection{Evaluation of the upscaling PAU-Net model}
Table \ref{table:unet_upscale_training_results} shows comprehensive results comparing the 4-level model variant against the 5-level model variant. The former achieved competitive results for the BraTS 2018 testing data and even superior results on the BraTS 2021 testing data. Interestingly, the 5-level model reached better metrics when regarding the performance results for the BraTS 2018 and BraTS 2021 training data, which, however, is not necessarily a benefit. On the BraTS 2018 training data, the 5-level model showed higher values on TC and WT and was only very little behind the 4-level model on ET. On the BraTS 2021 training data, both models showed very similar results. In summary, the 4-level model was superior on the test data, but slightly inferior on the training datasets. This was a strong indicator for an overfitting behaviour of the 5-level model, while the 4-level model was still generalizing well. The test results for the trained 4-level version of the model on the BraTS 2018 and the BraTS 2021 datasets are illustrated in Figure \ref{fig:unet_upscale_training_results}. These observations corroborate literature opinions \cite{Goodfellow2016}, \cite{Bishop2006} that less complex models are less sensitive to noise, enforce regularization effects and the stability of the model.  The reduced depth of the model's architecture  further allowed larger input size without exceeding the GPU memory of the available hardware. Table \ref{table:preact_level_comparsion} shows the number of parameters of both model versions, as well as the training speed and a maximum possible input data size for the given computation hardware.
%%%%%%%%%%%%%%%%%%%%%%%%%%%%%%%%%%%%%%%%
%%%%%%%%%%%%%%%%%%%%%%%%%%%%%%%%%%%%%%%%
\subsection{NET volume distributions}

\begin{figure}[!htbp]
  \centering
  \includegraphics[width=210pt]{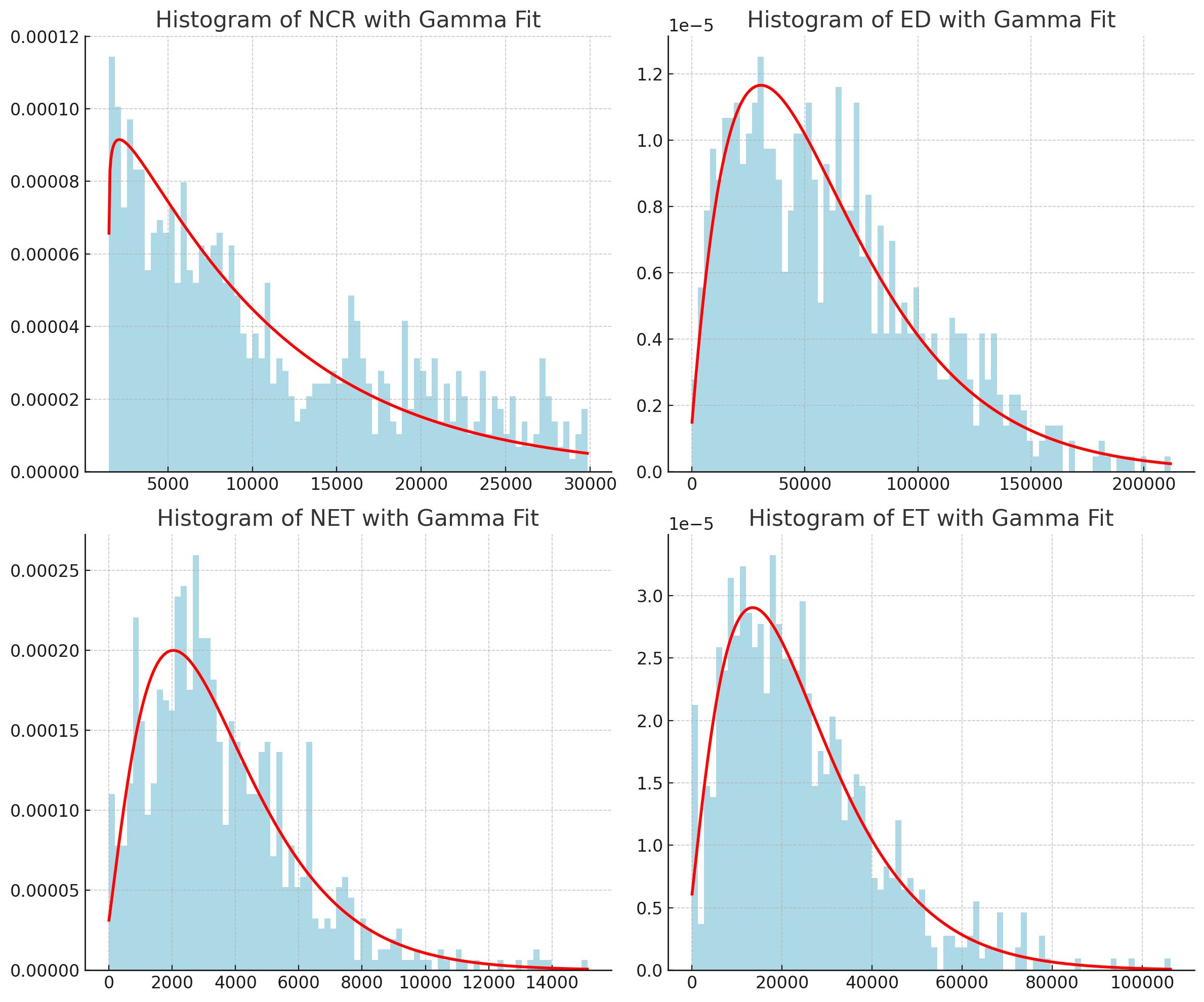}
  \caption{Gamma distribution fit to volume histograms}
  \label{fig:histo-fit}
\end{figure}

The distributions of the tumor compartment volumes of all four classes were right skewed and could be modeled by a gamma distribution. Figure \ref{fig:histo-fit} exhibits volume distributions fitted by a gamma distribution. The newly generated NET volumes showed distributions across the patient data in the BraTS 2021 dataset similar to the remaining classes.

%%%%%%%%%%%%%%%%%%%%%%%%%%%%%%%
%%%%%%%%%%%%%%%%%%%%%%%%%%%%%%%
\subsection{Analysis of variance for intensity masks}

The mean intensities of the NET - compartments showed significant differences to most of the other regions in the different modalities, i.e. we have

\begin{itemize}
    \item
    In T1: NET mean is significantly different to NCR and ED means.
    \item
    In T2: NET mean is significantly different to NCR and ET means.
    \item
    In T1C: NET mean is significantly different to NCR, ED and ET means.
    \item
    In Flair: NET mean is significantly different to NCR, ED and ET means.
\end{itemize}

Since the NET segment was placed within the edema compartment and was likely to be surrounded by it, the similar means of both in the T2 modality would make the detection difficult. The same held for the T1 modality, where NET and the ET compartments had very similar intensity values while also being located near or next to each other. This rendered T1C and Flair the two reasonable candidates for allocating the NET segments, although in practice this was still a demanding task and in many cases, the remaining tumor regions NCR, ED and ET were much easier to determine (see supplement for details).
%%%%%%%%%%%%%%%%%%%%%%%%%%%%%%%%%%%%%%%%%%%%%%%%
%%%%%%%%%%%%%%%%%%%%%%%%%%%%%%%%%%%%%%%%%%%%%%%%
\subsection{Unified 4-Label BraTS 18/21 Dataset}

The new large combined BraTS 2018/21 dataset encompassed a total number of $1536$ records together with the four labeled segments NCR, ED, NET and ET. The upscaling PAU-Net has again been trained on this larger dataset for retrieving better segmentation results. Besides the three original evaluation segments ET, TC and WT, two additional evaluation segments were introduced thereby: By adding NET as a fourth evaluation segment, the traditional evaluation segments ET, TC and WT could be compared more faithfully with reference models. The TC compartment was composed similar to the one in the original BraTS 2021 challenge, but excluding the NET regions, which originally were merged with the edema compartment. The TC compartment, including the NCR regions as well as the active ET region, is commonly also in the focus of a clinical context of treatment or resection. The NET compartment, on the other hand, is a frequently neglected compartment, often due to identification problems of the labeling experts. But due to the NET’s importance in medical treatment and tumor diagnosis, the TC compartment is increasingly suggested to be extended by the NET region. The newly assigned, extended TC compartment has thus been denoted TCN and included the NCR, the ET and the NET segments. This newly combined TCN compartment has been added as a fifth evaluation segment in the training of the large upscaling pre-activation U-Net model on the unified BraTS 2018/21 data. All evaluation segments and their composition based on isolated tumor compartments are listed in Table \ref{table:5comp}.

\begin{table}[!htb]
\caption{Unified BraTS 2018/21 evaluation segments as compositions of data labels. NET is added as a new evaluation segment as well as TCN, which extends the traditional TC labeled segment by the NET segment.}
\label{table:5comp}
\begin{scriptsize}
  \begin{center}
    \begin{tabular}{|c|c|c|c|}
    \hline
    \textbf{Eval. Segment} & \textbf{BraTS 2018/21 Data} & \textbf{BraTS 2018/21 Label} & \textbf{Description} \\
    \hline
    ET & ET & 4 & Enhancing Tumor\\
    TC & ET + NCR & 4,1 & Tumor Core\\
    WT & ET + NCR + NET + ED & 4,1,3,2 & Whole Tumor\\
    \hline
    NET & NET & 3 & Non-Enhancing Tumor\\
    TCN & ET + NCR + NET  & 4,1,3 & Tumor Core + Non-Enhancing Tumor \\
    \hline
    \end{tabular}
  \end{center}
\end{scriptsize}
\end{table}

\begin{table}[!htb]
 \caption{Dice-S\o{}rensen Coefficient test results obtained with unified 4-label BraTS 2018/21 data. Our mean values were computed from ET, TC and WT, not taking into account NET and TCN. }
\label{table:testres}
\begin{footnotesize}
  \begin{center}
    \begin{tabular}{|c|c|c|c|c|c|c|}
    \hline
    & \textbf{SegResNetVAE} & \textbf{TSC U-Net} & \textbf{nnU-Net} & \textbf{SegResNet} & \textbf{nnU-Net} & \textbf{PAU-Net} \\
    & 2018 \cite{Myronenko2018} & 2019 \cite{Jiang2020} & 2020 \cite{Isensee2020} & 2021 \cite{Siddiquee2021} & 2021 \cite{Luu2021} & \\
    \hline
    NET &  &  &  &  &  & 0.5320 \\
    TCN &  &  &  &  &  & 0.8846 \\ \hline
    ET   & 0.8145 & 0.8021 & 0.7945 & 0.8600 & 0.8823 & 0.8194\\
    TC   & 0.8596 & 0.8647 & 0.8524 & 0.8868 & 0.9235 & 0.8979\\
    WT  & 0.9042 & 0.9094 & 0.9119 & 0.9265 & 0.9383 & 0.9151\\
    {\bf Mean} & {\bf 0.8594} & {\bf 0.8587} & {\bf 0.8529} & {\bf 0.8911} & {\bf 0.9147} & {\bf 0.8775}\\
    \hline
    \end{tabular}
  \end{center}
\end{footnotesize}
\end{table}

The last column of Table \ref{table:testres} shows the performance results of our model as Dice-Sørensen coefficient, evaluated on the test data. On the ET compartment, the new model reached a Dice-score of $81.94 \%$, for the TC compartment $89.97 \%$ and for WT compartment $91.51 \%$. This resulted in a mean Dice-coefficient over the  ET, TC, WT segments of $87,75 \%$. Table \ref{table:testres} also shows performance results of the previously discussed BraTS winning models. Our upscaling PAU-Net model, which was trained on the unified BraTS 2018/21 dataset, achieved a competitive performance compared with the winning models. It outperformed the SegResNetVAE model \cite{Myronenko2018}, the TSC-U-Net \cite{Jiang2020} as well as the nnU-Net \cite{Isensee2020} in all three classical categories, while additionally and consistently providing the NET predictions. Furthermore, it is very competitive compared to the best performing models SegResNet \cite{Siddiquee2021} and nnU-Net \cite{Luu2021}. Concurrently, it offered two important advantages over existing models. On the one hand, it automatically outputs resolution enhanced segmentation masks serving a more detailed view onto the tumor’s structure. This can help radiologists and clinicians to visualize and understand the shape and extent of the tumor and it’s compartments. On the other hand, in addition to the traditional segments, it also predicts the important NET - compartments, which are hard to locate on the conventional radiology records. This can give new options in radiology assistance and treatment planning. A further aspect in analyzing and diagnosing the active tumor region is the extended TCN segment, which extends the classical tumor core TC by the relevant NET-compartment and can be a sophisticated complement in the hierarchical order of a glioma’s extent: $ET \subset TC \subset TCN \subset WT$.  The overall performance score of TCN, using the proposed model on the unified BraTS 2018/21 dataset, was $88.46 \%$, which is very comparable to the one of TC, which was $89.79 \%$. The DSC score of NET on the given dataset was $53.20 \%$, which is low compared to the remaining segments. This is due to the subtle nature of the NET-compartments, which are showing very weak contrasts on conventional MRI sequences. However, one also has to be aware of common shortcomings of canonical metrics, an aspect that needs further investigations in future work \cite{Reineke2023a}, \cite{Reineke2024}. The lack of curated NET tumor compartment data and the practical limitations in representative reconstruction from given data are also contributing to inhomogeneous NET data in the constructed dataset. This necessarily leads to an overall upper bound performance in NET identification, given a proper regularization behaviour and robustness of the model.

%%%%%%%%%%%%%%%%%%%%%%%%%%%%
%\bibliographystyle{natbib}
\bibliographystyle{plain}
\bibliography{BTS}
%%%%%%%%%%%%%%%%%%%%%%%%%%%%%%%%%%%%
%%%%%%%%%%%%%%%%%%%%%%%%%%%%%%%%%%%%
%%%%%%%%%%%%%%%%%%%%%%%%%%%%%%%%%%%%
\end{document}